\documentclass[a4paper, 10pt, conference]{ieeeconf}      

\IEEEoverridecommandlockouts                              

\usepackage{graphicx} 

\title{\LARGE \bf
Enhancing Worker Safety in Harbors Using Quadruped Robots
}

\author{ Zoe Betta $^{1}$, Davide Corongiu $^{2}$, Carmine Tommaso Recchiuto $^{1}$ and Antonio Sgorbissa $^{1}$
\thanks{*This work was not supported by any organization}
\thanks{$^{1}$ Rice Lab, University of Genova, Via Opera Pia 13, 16145, Genova, Italy and
RAISE Ecosystem, Genova, Italy. Contact author:
        {\tt\small zoe.betta@edu.unige.it}}%
\thanks{$^{2}$ Autorità di Sistema Portuale del Mar Ligure Occidentale, Via della Mercanzia, 2, 16124 Genova, Italy
}
}

\begin{document}

\maketitle
\thispagestyle{empty}
\pagestyle{empty}

\begin{abstract}
Infrastructure inspection is becoming increasingly relevant in the field of robotics due to its significant impact on ensuring workers' safety. The harbor environment presents various challenges in designing a robotic solution for inspection, given the complexity of daily operations. This work introduces an initial phase to identify critical areas within the port environment. Following this, a preliminary solution using a quadruped robot for inspecting these critical areas is analyzed.

\end{abstract}

\section{INTRODUCTION}
In Italy, in 2023 there were a total of 590,215 occupational injuries, of which 1,147 resulted in death \cite{inail}.
To mitigate the impact of accidents, researchers 
are deploying robotic solutions to conduct autonomous infrastructure inspections \cite{ insp2, insp3}. 
 To complete industrial inspections, different types of robots have been deployed: wheeled robots, aerial drones, legged robots, and custom-designed robots. 
Some robot designs depend on the specific solution and context, such as pipe inspection \cite{pipe1, pipe2}
, making them difficult to adapt for other applications. Among other options, quadruped robots are often recommended for their agility and ability to adapt to different terrains \cite{quad1, quad2}. For example, in 2022, the Spot robot was deployed at Austria’s largest power plant to enhance worker safety \cite{austria}. In this scenario, the robot followed pre-determined routes throughout the plant and reported any detected safety concerns.

\begin{figure}[ht]
    \centering
    \includegraphics[width=0.6\columnwidth]{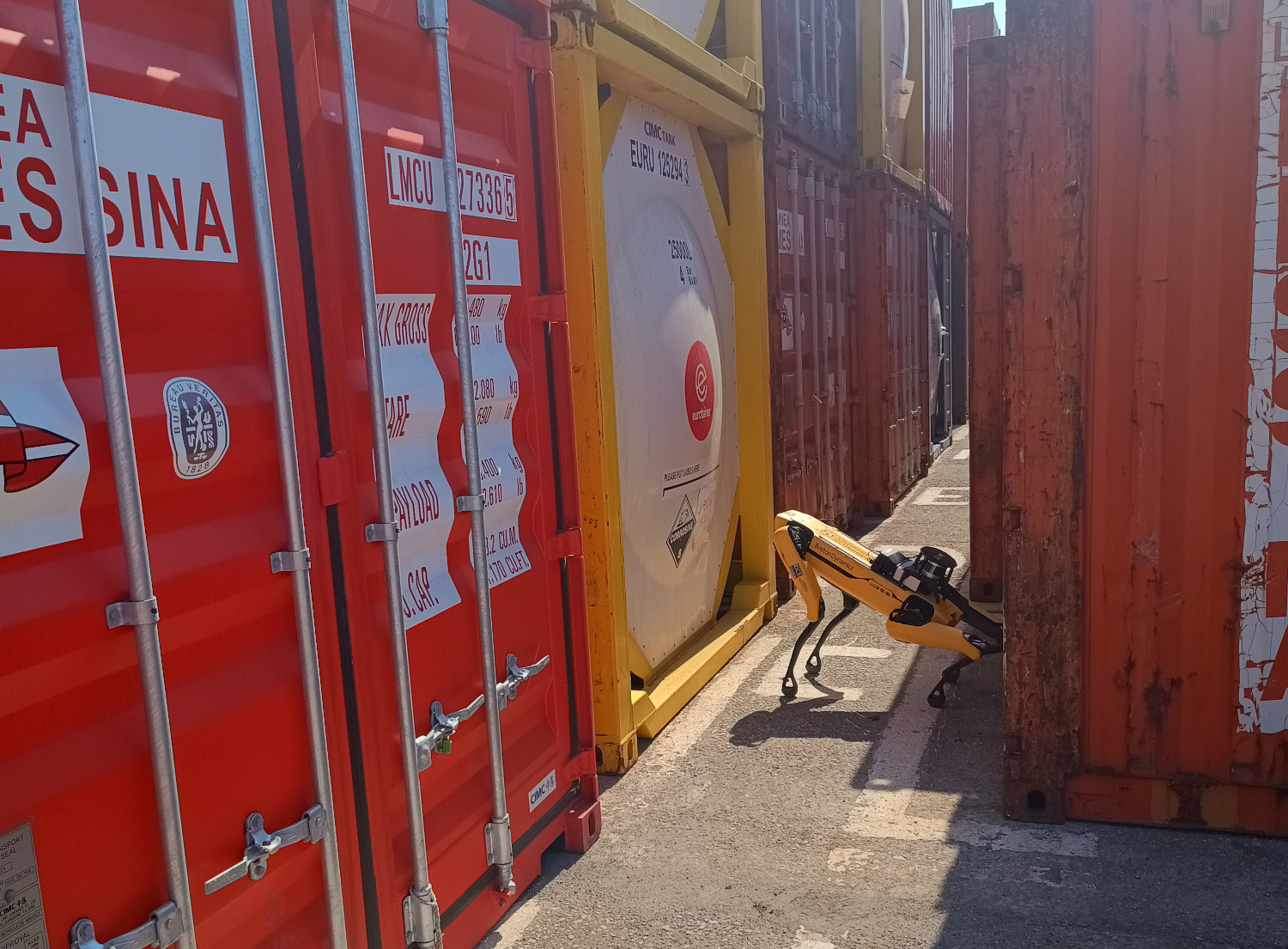}
    \caption{Spot robot inspecting a container at the Harbor of Genova}
    \label{fig:spot}
    \vspace*{-5pt}
\end{figure}

In this work, we aim to develop a solution for a quadruped robot to monitor a harbor environment and enhance workers' safety. Figure \ref{fig:spot} shows the Spot robot navigating between containers while observing the labels on their sides. This image was taken during a demonstration of the robot's motion capabilities under teleoperation.

Genova's harbor, the largest port in Italy, has a terrain extension of 7,000,000 m$^2$. Due to its size and outdoor nature,
a drone might appear to be an effective solution. Drones can move quickly from one place to another, 
and providing a different perspective from the one available to workers. On the other hand, drones have limited battery life and require special permits, which are not always possible to obtain in all harbors—for example, in Genoa’s harbor, due to its close proximity to the airport and its flight cone. In addition, drones might be dangerous if used in areas where workers are present or in very confined spaces where collisions are likely. 
Quadruped robots, on the other hand, have a longer battery life and are designed to operate in close proximity to workers without requiring special permits. Additionally, they can carry heavier loads and move through small spaces while also climbing obstacles and stairs. However, they are slower in moving from one area of the harbor to another and have some mobility limitations. Quadruped robots can operate in areas that are, in principle, accessible to humans as well. The reason for using a quadruped robot is not the inaccessibility of the location but rather to enhance safety and expose only the robot to potential dangers.
Wheeled robot, even if able to carry heavier weights and maximizing battery life, are too limited in mobility, they require almost flat surfaces, for this reason are not considered as possible solutions. 

The contributions of this work are: (1) identifying critical areas and scenarios where security needs improvement in the harbor environment and (2) assessing which of these scenarios could benefit from the deployment of a sensorized quadruped robot.
To this end, we conducted a series of semi-structured interviews with harbor workers representing different terminals and roles. The insights gathered were then analyzed as a preliminary step to design a quadruped robot aimed at enhancing workers' safety in the port environment. 
\begin{table*}[h]
\centering
\caption{Emerged Problems}
\label{tab:problem}
\begin{tabular}{l|l|l|l|l}
\hline
 & Problems                                                       & Location     & Priority & Check repetition\\ \hline
1 & Custom inspection                                              & Storage Yard &    Low   & When needed   \\
2 & Chemical leaks in the hazardous good park                      & Storage Yard & High  & Repeated   \\
3 & Radioactive materials in hazardous good park                   & Storage Yard &   High  & Repeated      \\
4 & Fires at the hazardous goods park                              & Storage Yard &    High  & Repeated     \\
5 & Incorrect piling of containers                                 & Storage Yard &    Medium  & Repeated    \\
6 & Self-combustion of bulk storage (e.g. coal)                    & Storage Yard &      High  & Repeated   \\
7 & Unauthorized people in restricted access areas                 & Storage Yard &   Low   & Repeated    \\
8 & Wrong application of train seals                               & Storage Yard &     Low  & Repeated   \\
9 & Poor air quality in ships' holds                               & Ships' hold  &       High & When needed   \\
10 & Unstable cargo due to weather adversities or incorrect loading & Ships' hold  &    High   & When needed    \\
11 & Poor air quality in naval repairs                              & Ships' hold  &   High   & When needed     \\
12 & Radioactive materials in naval repairs                         & Ships' hold  &  High   & When needed      \\
13 & X-ray inspection of cranes                                     & Cranes       &    Low  & Repeated    \\
14 & Inspection of cranes with penetrating liquids                  & Cranes      &    Low  & Repeated     \\
15 & The Twist-mechanisms of containers get blocked                & Storage Yard       &      Medium  &  When needed
\end{tabular}
\vspace*{-10pt}
\end{table*}

In recent years, inspection tasks in civil environments have increasingly incorporated robotic solutions. The work of Lattanzi et al. \cite{inspectionreview} is a review focusing on 
several application fields such as buildings, bridges, and tunnels. However, the harbor environment is never mentioned, and there is little attention given to specific settings with its unique challenges.  
The work of Choi et al. \cite{port3} and Tani et al. \cite{port1} focuses on the development of underwater robots for inspecting port structures below the water surface. While this is a crucial aspect of inspection, it is important to recognize that most workers operate on the land side of the harbor, highlighting the need for monitoring land-based operations as well. Furthermore, in both studies, the potential users of the robot are not mentioned. Engaging with workers is essential to ensure the system's acceptance and effective integration into their operations.
The work of Santos et al. \cite{coopinspection} explores a solution for harbor inspection using both drones and underwater robots. The focus of this work is the development of a strategy for a drone to autonomously land on a water-based robot after completing its inspection task. While drones are useful for inspecting large harbor areas, they are less suitable for conducting detailed inspections of smaller areas, where closer proximity and higher-resolution data acquisition are required. Additionally, this study does not consider workers' opinions on the proposed solution.

This paper is structured as follows: Section \ref{sec:met} describes the methodology. Sections \ref{sec:results} and \ref{sec:discussion} present and discuss the results. Finally, Section \ref{sec:conclusion} provides the conclusions.

\section{METHODOLOGY}
\label{sec:met}

The first step 
involved meeting with workers in various areas of Genova's harbor. We visited three terminals: Terminal IMT, Terminal C. Steinweg - GMT SRL, and Terminal Rinfuse Genova.
Each terminal manages the loading and unloading of ships, but they differ in the type of goods handled. Terminal IMT specializes in handling containerized goods, and provides storage for hazardous materials. Terminal C. Steinweg - GMT SRL focuses on aluminum products, such as slabs and coils. Terminal Rinfuse Genova handles bulk goods, storing them in open-air containers.

We interviewed a total of 10 people: 5 from Terminal IMT, 3 from Terminal C. Steinweg - GMT SRL, and 2 from Terminal Rinfuse Genova. At each Terminal, we ensured that the interviews included individuals with different backgrounds and roles. Specifically, we required to speak with someone responsible for coordinating workers, the safety manager, and several workers. Additionally, during interviews, the Head of the health and safety inspection departments, from Port Authority, was always present. He manages safety concerns in the harbor environment while maintaining a mediator role. The interview was structured in three phases:
\begin{itemize}
    \item Preliminary phase, in which we aimed to gather information about the workflow and possible critical areas in each terminal.
    \item Demonstration phase, in which we showed the capabilities of the Spot robot from Boston Dynamics.
    \item Design phase, in which we collaboratively envisioned an application with workers using the previously demonstrated robot and gathered insights into their perceptions of the robot.
\end{itemize}

During the interview, we decided not to describe our previous research with Spot to allow participants to freely imagine the application, without any external influence. We revealed our previous works only at the end of the meeting after gathering all of the information. The interviews were recorded and later transcribed verbatim for analysis. 
We also planned, but have not yet completed, an interview and demonstration on a Roll-On Roll-Off (Ro-Ro) ship. Ro-Ro ships are commercial vessels specialized in transporting wheeled cargo. The peculiarity of these ships is that the cargo is driven onto the ship with trucks rather than being loaded using cranes. The driver, after loading the cargo, usually drives off the ship with the power unit, leaving only the cargo to be transported. At the destination port, another driver will load the cargo with their tractor unit and drive it off the ship. Scheduling this interview has been especially challenging because the ships remain in the harbor for only a limited time, and the interview must not interfere with the ship’s normal operations.

\section{RESULTS}
\label{sec:results}

We analyzed the interviews to identify the occurring problems.
Recurrent themes surfaced from the analysis of the interviews, as reported in Table \ref{tab:problem}. In the first column of the table, we include an identifier to facilitate association with the proposed solutions in Table \ref{tab:design}. In the second column, we provide a brief description of each problem.

\begin{figure}[ht]
    \centering
    \includegraphics[width=0.6\columnwidth]{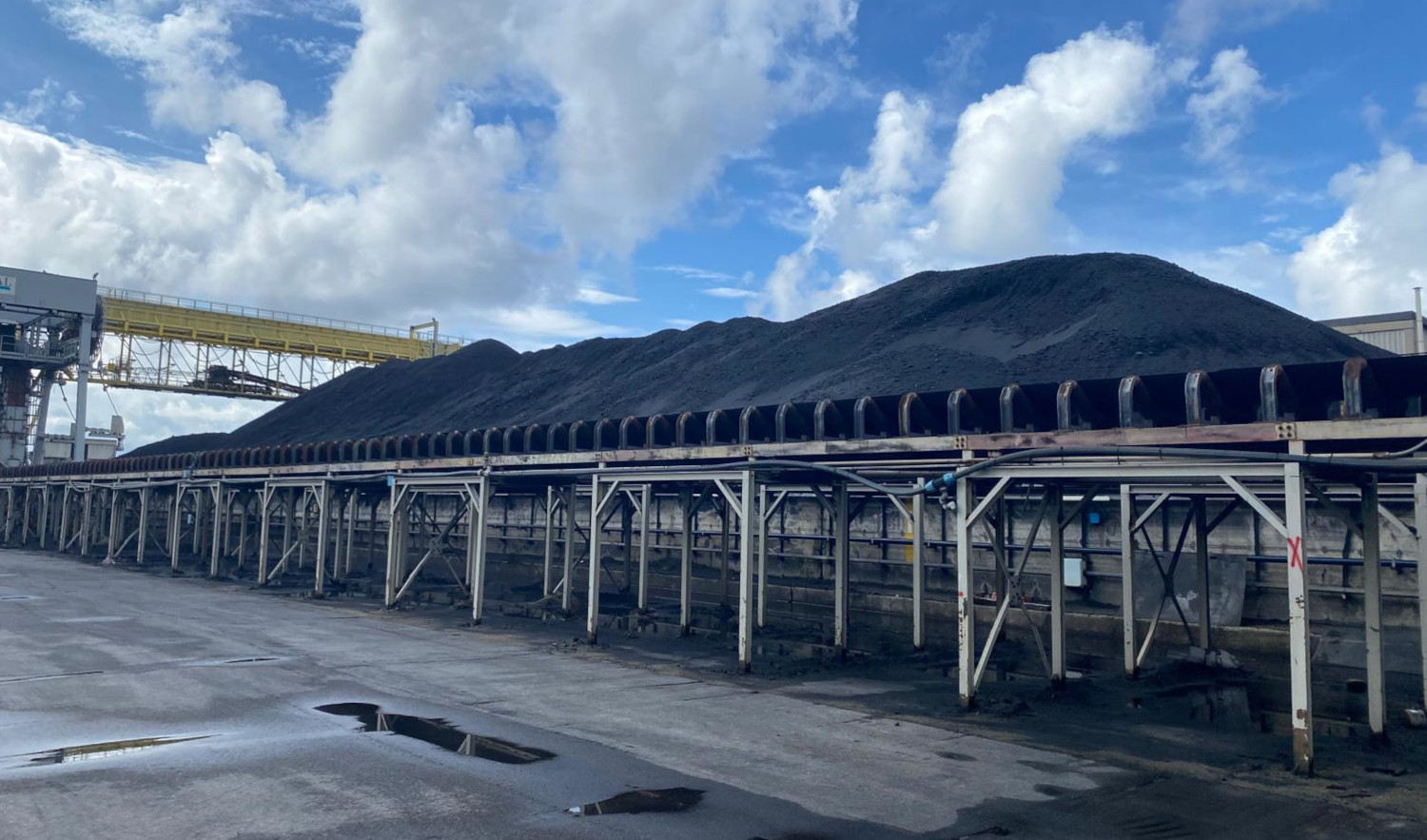}
    \caption{Terminal Rinfuse Genova storing bulked goods (coal).}
\vspace*{-5pt}
    \label{fig:bulk}
\end{figure}

First, it emerges that critical areas can be classified into three categories based on specific needs and the locations where the robot might operate, as shown in the third column of Table \ref{tab:problem}: storage yards (60\%), ships' hold (27\%), and crane (13\%).
The storage yard is present in several terminals and can hold both containers, as shown in Fig. \ref{fig:spot}, and bulk goods, as shown in Fig. \ref{fig:bulk}. It stores all materials or goods received at the port that need to be kept before being shipped to other destinations, either via other ships or by land.
In some scenarios, inspections must be conducted onboard ships, typically in the ship's hold. This is necessary because, when a ship arrives at the harbor for unloading, workers are often unaware of the safety conditions inside (e.g., due to adverse weather conditions in the preceding days). Similarly, when workers need to operate in confined spaces for naval repairs, they also lack information about the safety conditions. 
Finally, during the interviews, while collecting information on the safety challenges faced by workers, the topic of aerial operations emerged. This category includes all operations requiring an operator to complete tasks at an elevated altitude using cranes.


On the fourth column we reported the priority given to the different problems. Depending on the time spent on the problem during the interview, and the words used to describe the urgency of finding a solution we classified each problem as high, medium, or low priority. 
Column five indicates the expected frequency of inspections. This categorization differentiates between inspections that are conducted periodically—such as on a daily basis—and those that must be performed at specific times following an event, such as when a ship arrives at the harbor.
\begin{table*}[bt]
\centering
\caption{Possible Solutions}
\label{tab:design}
\begin{tabular}{l|l|l|l}
\hline
 & Design solutions                                               & Preferred robot  & Autonomy Level        \\ \hline
1 & Moving inside the container to verify the contents                                             & Drone     & Teleoperated               \\
2 & Inspection to verify the presence of chemicals near the containers                      & Quadruped robot   & Autonomous       \\
3 & Inspection to verify the presence of radioactions in the hazardous goods park                 & Quadruped robot   & Autonomous       \\
4 & Inspection to verify the presence of fires in the hazardous goods park                             & Quadruped robot   & Autonomous       \\
5 & Visual inspection to verify correct piling of containers                                & Quadruped robot     & Autonomous     \\
6 & Inspection to monitor the temperature of the bulk goods                    & Quadruped robot    & Autonomous      \\
7 & Inspection to verify the presence of people in restricted areas                & Quadruped robot/drone & Autonomous \\
8 & Visual inspection of train seals                             & Quadruped robot/drone & Autonomous\\
9 & Inspection to verify the percentage of CO$_2$ and O$_2$ in the ships' holds                              & Quadruped robot  & Mixed Autonomy        \\
10 & 3D map of the cargo to verify its positioning & Quadruped robot   & Mixed Autonomy       \\
11 & Inspection to verify the percentage of CO$_2$ and O$_2$ during naval repairs                              & Quadruped robot   & Mixed Autonomy       \\
12 & Inspection to verify the presence of radioactions during naval repair                       & Quadruped robot   & Mixed Autonomy       \\
13 & X-Ray sensors on drones                                     & Drone      & Autonomous              \\
14 & Drones equipped with penetrating liquids and high resolution cameras                  & Drone    & Autonomous                \\
15 & Visual inspection of the Twist mechanism               & Drone    & Teleoperated              
\end{tabular}
\vspace*{-10pt}
\end{table*}
We then proceeded with the next phase, during which we 
reviewed the interviews to understand the possible solutions that were mentioned for each problem, 
as shown in Table \ref{tab:design}.

In the first column, we report the identifier 
associated with the corresponding problem. 
The second column contains a description of the possible solution. 
The third column lists the preferred robot mentioned by the workers. Although the focus was on applications using a quadruped robot, workers also suggested some use cases involving drones.
Finally, in the fourth column, we indicate the level of autonomy expected by the workers. Based on their descriptions of the operation, we classified the applications into three categories: teleoperation, mixed autonomy, or autonomous.

\section{DISCUSSION}
\label{sec:discussion}

We first analyze the tasks that a quadruped robot can perform in the storage yard. 
For inspections in the hazardous goods area—such as monitoring potential chemical leaks (2), radiation (3), and fires (4)—a quadruped robot is a suitable choice due to its ability to climb obstacles and navigate around people. It can move close to ground-level features for inspection, similar to how a human would, and maneuver in tight spaces to approach containers, enhancing the precision of leak detection near potential sources. To carry out these tasks, the robot must be equipped with 
sensors, including gas sensors for leak detection, radiometric sensors for radiation monitoring, and thermal cameras for fire detection.



For the task of monitoring self-combustion in the bulk storage area (6), the workers we interviewed suggested that a quadruped robot might be a better choice than a drone due to its ability to move directly on top of the pile. This capability allows the robot, when equipped with thermal cameras, to detect temperature variations more precisely. Additionally, a quadruped robot equipped with high-resolution cameras and an IMU sensor can more effectively identify improper container stacking by assessing both the relative positioning of the corners and the vertical alignment of the pile with respect to gravity (5). This task is more challenging for a drone, as it is influenced by weather and wind conditions.
To monitor access to restricted areas (7) and ensure the correct application of train seals (8), both drones and quadruped robots are appropriate choices, when equipped with high resolution cameras. Drones move faster and gather more information, while quadruped robots are more precise in detecting people or classifying seals.

As a general consideration, all tasks from 2 to 8 must be performed repeatedly. Due to the dynamic nature of the harbor, it is impossible to predict the location of obstacles in advance. Consequently, programming a fixed route for the robot is not feasible. Instead, flexibility requires functionalities typical of fully autonomous robots, such as exploration, mapping, planning, and the detection of relevant features. 



For the ships' hold applications, the robot must be equipped with various sensors, such as radioactive sensors, gas sensors, and a 3D LiDAR.

For bulk goods inspection in ships' holds (9) or Ro-Ro inspections upon a ship's arrival (10), 
LiDAR is used to generate a precise 3D representation of the hold, allowing for the assessment of cargo stability, such as detecting broken chains or tipped-over containers. In this case, a quadruped robot provides a significant advantage in gathering information about the entire hold, as it can move under the cargo and navigate over small obstacles.
For the naval repair scenario (11 and 12), only radioactive sensors and gas sensors are required. In this case, the robot needs to be deployed in hard-to-reach locations before workers arrive to begin their tasks.
For both the ship unloading and naval repair scenarios, it is crucial to consider the robot's agility and dimensions. The robot must be able to navigate narrow spaces.

Since these tasks must be performed as needed—either as soon as the ship arrives for unloading or before starting a series of repairs—the robot could, in principle, be teleoperated. For Ro-Ro ships, for example, the robot could be teleoperated to the end of the hold to collect data. However, the crowded environment may cause disturbances.
For this reason, it would be best to implement a mixed-autonomy behaviour to ensure the robot's continuous operation. The robot could be teleoperated when the connection is stable and switch to autonomous behaviors when the connection becomes unreliable.


Other interesting applications emerged, even though they 
do not involve quadruped robots. For customs inspections (1), the goal is to deploy an agile robot capable of moving inside partially empty containers to verify their contents. 
A drone could fly inside the container without exerting pressure on potentially fragile items, making it a more suitable option. 
Also, the workers mentioned the need to improve crane inspections using robotic solutions. 
Currently, inspections of cranes are carried out by workers suspended in the air who apply penetrating liquids to the crane to check for cracks or perform X-ray scans of the interested areas. 
The workers 
suggested a potential solution using drones (13 and 14). 
The task regarding the failure of the twist-lock mechanism will not be discussed here, as its complexity would require more information to hypothesize a solution (15). 


During the interview, while the majority of the workers were enthusiastic about adopting robotics solutions to improve safety in their workplace, some were hesitant. A few workers expressed concerns about the potential loss of jobs. 
Additionally, some mentioned that robots might not be as precise as people in completing certain tasks. 
Developing solutions in collaboration with workers is crucial to understanding their concerns and addressing them properly during the design phase, 
so that they do not perceive robots as a replacement for human labor but rather an instrument they can use to improve their safety. 

\section{CONCLUSION}
\label{sec:conclusion}
This work presents the results of semi-structured interviews with workers in various roles at the Port of Genova. The interviews focus on identifying critical areas for worker safety within the harbor. Each problem has been analyzed to explore possible solutions for deploying a quadruped robot. Engaging with end users is essential to tailor the robotic solution to their requirements and improve its acceptance.

We first examined workers' needs, identifying the areas where these needs arise, their priority, and their frequency. Then, we analyzed possible solutions, determining whether an aerial drone or a quadruped robot is more suitable and specifying the required level of autonomy.


This work 
will serve as a foundation for future developments aimed at implementing and testing the robotic solution in the harbor environment.

\addtolength{\textheight}{-12cm}   




\section*{ACKNOWLEDGMENT}
This work was carried out within the framework of the project "RAISE - Robotics and AI for Socio-economic Empowerment” Spoke 4 and has been supported by European Union – NextGenerationEU. We thank the staff of Autorità di Sistema Portuale del Mar Ligure Occidentale, Terminal IMT, Terminal C. Steinweg - GMT SRL, and Terminal Rinfuse Genova for their help.

\bibliographystyle{IEEEtran.bst}
\bibliography{bib.bib}

@electronic{inail,
  title         = "Banca dati statistica",
  url           ="https://www.inail.it/portale/it/attivita-e-servizi/dati-e-statistiche/banca-dati-statistica.html",
  year          = "2024",
  key           = "hold"
}

@electronic{austria,
  title         = "Spot Makes Austria’s Largest Power Plant Safer",
  url           ="https://bostondynamics.com/case-studies/spot-makes-austrias-largest-power-plant-safer/",
  year          = "2024",
  key           = "hold"
}

@INPROCEEDINGS{port1,
  author={Tani, Simone and Ruscio, Francesco and Costanzi, Riccardo},
  booktitle={OCEANS 2024 - Halifax}, 
  title={Preliminary Online Validation of a Visual-Acoustic-based Framework for Autonomous Underwater Structure Inspections}, 
  year={2024},
  volume={},
  number={},
  pages={1-8},
  doi={10.1109/OCEANS55160.2024.10753951}}

@INPROCEEDINGS{port3,
  author={Choi, Jinwoo and Lee, Yeongjun and Kim, Taejin and Jung, Jongdae and Choi, Hyun-Taek},
  booktitle={Proc. UT'17)}, 
  title={Development of a ROV for visual inspection of harbor structures}, 
  year={2017},
  volume={},
  number={},
  pages={1-4},
  doi={10.1109/UT.2017.7890285}}

@article{inspectionreview,
author = {David Lattanzi  and Gregory Miller },
title = {Review of Robotic Infrastructure Inspection Systems},
journal = {Journal of Infrastructure Systems},
volume = {23},
number = {3},
pages = {04017004},
year = {2017},
doi = {10.1061/(ASCE)IS.1943-555X.0000353}

}

@ARTICLE{coopinspection,
	author = {Santos, Matheus C. and Bartlett, Ben and Schneider, Vincent E. and Brádaigh, Fiachra O. and Blanck, Benjamin and Santos, Phillipe C. and Trslic, Petar and Riordan, James and Dooly, Gerard},
	title = {Cooperative Unmanned Aerial and Surface Vehicles for Extended Coverage in Maritime Environments},
	year = {2024},
	journal = {IEEE Access},
	volume = {12},
	pages = {9206 – 9219},
}

@ARTICLE{insp2,
  author={Fischer, Georg K. J. and Bergau, Max and Gómez-Rosal, D. Adriana and Wachaja, Andreas and Graeter, Johannes and Odenweller, Matthias and Piechottka, Uwe and Höflinger, Fabian and Gosala, Nikhil and Wetzel, Niklas and Büscher, Daniel and Valada, Abhinav and Burgard, Wolfram},
  journal={IEEE Sensors Journal}, 
  title={Evaluation of a Smart Mobile Robotic System for Industrial Plant Inspection and Supervision}, 
  year={2024},
  volume={24},
  number={12},
  pages={19684-19697},
}

@inproceedings{insp3,
author = {Xi, Xiangji and Zhao, Jian and Wang, Hui and Yan, Tianyu and Liu, Jun},
title = {Research on Recognition Algorithm of Power Facility Inspection Robot Based on Computer Vision},
year = {2024},
address = {New York, NY, USA},
doi = {10.1145/3689236.3689326},
booktitle = {Proc. ICCSIE '24},
pages = {446–453},
}

@ARTICLE{pipe1,
  author={Khan, Muhammad Bilal and Chuthong, Thirawat and Danh Do, Cao and Thor, Mathias and Billeschou, Peter and Larsen, Jørgen Christian and Manoonpong, Poramate},
  journal={IEEE Access}, 
  title={iCrawl: An Inchworm-Inspired Crawling Robot}, 
  year={2020},
  volume={8},
  number={},
  pages={200655-200668},
  doi={10.1109/ACCESS.2020.3035871}}

@INPROCEEDINGS{pipe2,
  author={Nivethika, S Deepa and Nivethetha, T and Priyadharshini, P and Nithyasri, V T and SenthilPandian, M and Sivaprasad, R},
  booktitle={Proc. ICPECTS'22}, 
  title={Design and Development of Pipe inspection Snake Locomotion Robot}, 
  year={2022},
  volume={},
  number={},
  pages={1-5},
}

@InProceedings{quad1,
author="Stefaniak, Pawel
and Anufriiev, Sergii",

title="Method of Defining Diagnostic Features to Monitor the Condition of the Belt Conveyor Gearbox with the Use of the Legged Inspection Robot",
booktitle="Intelligent Information and Database Systems",
year="2020",
publisher="Springer Singapore",
address="Singapore",
pages="158--167",
isbn="978-981-15-3380-8"
}

@INPROCEEDINGS{quad2,
  author={Parkinson, Brian and Wolf, Adám and Galambos, Péter and Széll, Károly},
  booktitle={Proc. INES'23)}, 
  title={Assessment of the Utilization of Quadruped Robots in Pharmaceutical Research and Development Laboratories}, 
  year={2023},
  volume={},
  number={},
  pages={221-228},
 }


\end{document}